\documentclass[runningheads]{llncs}
 
\usepackage{eccv}

\usepackage{eccvabbrv}

\usepackage{graphicx}
\usepackage{booktabs}

\usepackage[accsupp]{axessibility}  

\usepackage{hyperref}

\usepackage{orcidlink}
 
\usepackage{hyperref}
\usepackage{url}
\usepackage[utf8]{inputenc}
\usepackage[ruled,vlined,linesnumbered]{algorithm2e}
\usepackage{graphicx}
\usepackage{wrapfig}
\usepackage{enumitem}
\setlist{nosep}
\usepackage{longtable}
\usepackage{booktabs}
\usepackage{array}
\usepackage{geometry}
\usepackage{amsmath,amssymb}
\usepackage{listings}
\usepackage{xcolor}
\usepackage{multirow}
\usepackage{caption}
\usepackage{booktabs}
\usepackage{varwidth}
\usepackage{algpseudocode}
\setlist{nosep}
\usepackage{url}
\usepackage{floatflt}
\usepackage{float}
\newcolumntype{P}[1]{>{\centering\arraybackslash}p{#1}}
\usepackage{listings}
\usepackage[table]{xcolor}

\definecolor{goodgreen}{RGB}{0,150,0}
\definecolor{badred}{RGB}{180,0,0}

\lstdefinelanguage{yaml}{
  keywords={true,false,null,y,n},
  keywordstyle=\color{blue}\bfseries,
  sensitive=false,
  comment=[l]{\#},
  morecomment=[l]{\#},
  commentstyle=\color{gray}\ttfamily,
  stringstyle=\color{red}\ttfamily,
  moredelim=[s][\color{black}]{:}{\ },
  morestring=[b]",
  morestring=[b]'
}
\lstdefinelanguage{prompt}{
  morecomment=[l]{\#},
  commentstyle=\color{gray},
}

\lstdefinelanguage{json}{
    basicstyle=\ttfamily\small,
    numbers=left,
    numberstyle=\tiny,
    stepnumber=1,
    numbersep=5pt,
    showstringspaces=false,
    breaklines=true,
    frame=lines,
    literate=
     *{0}{{{\color{blue}0}}}{1}
      {1}{{{\color{blue}1}}}{1}
      {2}{{{\color{blue}2}}}{1}
      {3}{{{\color{blue}3}}}{1}
      {4}{{{\color{blue}4}}}{1}
      {5}{{{\color{blue}5}}}{1}
      {6}{{{\color{blue}6}}}{1}
      {7}{{{\color{blue}7}}}{1}
      {8}{{{\color{blue}8}}}{1}
      {9}{{{\color{blue}9}}}{1}
      {:}{{{\color{red}{:}}}}{1}
      {,}{{{\color{red}{,}}}}{1}
      {"}{{{\color{goodgreen}{"}}}}{1}
}

\newcommand{\system}{\textbf{\textsc{Remsa}}\xspace}
\newcommand{\fmd}{\textsc{RS-FMD}\xspace}
\newcommand{\stitle}[1]{\vspace{0.5ex}\noindent{\bf #1}}
\newcommand{\etitle}[1]{\vspace{1ex}\noindent{\underline{\em #1}}}

\begin{document}

\title{\system: Foundation Model Selection for Remote Sensing via a Constraint-Aware Agent} 


\author{Binger Chen\inst{1}\orcidlink{0009-0007-7153-6100}\thanks{Work done while the author was affiliated with Technische Universität Berlin \& BIFOLD.} \and
Tacettin Emre Bök\inst{2}\textsuperscript{*} \and
Behnood Rasti\inst{3}
\and
Volker Markl\inst{3}
\and
Begüm Demir\inst{3}}

\authorrunning{B.~Chen et al.}

\institute{Humboldt-Universität zu Berlin, Berlin, Germany\\
\email{binger.chen@hu-berlin.de}\and
5U AI, Munich, Germany\\
\email{tacettinemre1@gmail.com}\and
Technische Universität Berlin \& BIFOLD, Berlin, Germany\\
\email{behnood.rasti@gmail.com, \{volker.markl,demir\}@tu-berlin.de}}

\maketitle

\begin{abstract}
Foundation Models~(FMs) are increasingly integrated into remote sensing~(RS) pipelines for applications such as environmental monitoring, disaster assessment, and land-use mapping. 
These models include unimodal vision encoders trained in a single data modality and multimodal architectures trained in multiple sensor modalities, such as synthetic aperture radar~(SAR), multispectral, and hyperspectral imagery, or jointly in image-text pairs in vision-language settings.
FMs are adapted to diverse perception tasks, such as semantic segmentation, image classification, change detection, and visual question answering, depending on their pretraining objectives and architectural design. However, selecting the most suitable remote sensing foundation model~(RSFM) for a specific task remains challenging due to scattered documentation, heterogeneous formats, and complex deployment constraints. 
To address this, we first introduce the RSFM Database~(\fmd), the first structured and schema-guided resource covering over 160 RSFMs trained on various data modalities, spanning different spatial, spectral, and temporal resolutions, considering different learning paradigms.
Built upon \fmd, we further present \system~(\underline{\textbf{Re}}mote-sensing \underline{\textbf{M}}odel \underline{\textbf{S}}election \underline{\textbf{A}}gent), a constraint-aware agent that enables automated RSFM selection from natural language queries. 
\system combines structured FM metadata grounding with task-aware orchestration for retrieval, clarification, ranking, and explanation. In detail, it interprets user input, clarifies missing constraints, ranks models via in-context learning, and provides transparent justifications.
Our system supports various RS tasks and data modalities, enabling personalized, reproducible, and efficient FM selection. 
To evaluate \system, we construct a benchmark of 100 expert-verified RS query scenarios. 
Each query is evaluated across 4 systems and 3 LLM backbones, with the top-3 selected models assessed by domain experts using a fixed suitability rubric. 
This results in \textbf{3,000} expert-scored task--system--model configurations under our rubric-based expert evaluation protocol.
\system outperforms multiple baselines, including naive agent-based method, dense retrieval, and unstructured retrieval augmented generation approaches, showing its practical utility in real decision-making applications. \system operates entirely on publicly available metadata of open source RSFMs, without accessing private or sensitive data. Our code and data are publicly available at: \url{https://github.com/be-chen/REMSA}.
\end{abstract}

\section{Introduction}


With the growing availability of remote sensing~(RS) missions and their onboard sensors~(\eg, Sentinel-2~\cite{drusch2012sentinel}, Sentinel-1~\cite{torres2012gmes}, EnMAP~\cite{enmap}), RS plays an increasingly important role in many applications such as agriculture, disaster response, urban development, and biodiversity monitoring. These applications increasingly rely on foundation models~(FMs) that can generalize across various RS data modalities with different spatial, spectral, and temporal resolutions, geospatial extents and applications, while being transferable and effective even with limited labeled data in downstream RS analysis tasks.
Recently, numerous FMs have emerged in the RS domain, offering powerful capabilities for modeling complex RS data. These models include vision-only encoders trained on single or multiple RS data modalities~(\eg, MMEarth~\cite{mmearth}, OmniSat~\cite{omnisat}, MA3E~\cite{ma3e}) and vision–language models~(VLMs) trained with paired RS data modalities and text~(\eg, GeoText~\cite{geotext}, LHRS-Bot~\cite{lhrsbot}). These models are pretrained on large-scale and heterogeneous RS datasets encompassing a diverse range of sensor modalities, including RGB, multispectral, hyperspectral, synthetic aperture radar~(SAR), and light detection and ranging~(LiDAR), across multiple spatial and temporal resolutions.
Each FM exhibits its strengths in distinct downstream RS tasks, such as classification, object detection, change detection, captioning, and visual question answering~(VQA). 
For instance, in practical RS workflows, change detection typically relies on multitemporal SAR or optical data inputs, while fine-grained land cover mapping often benefits from high-resolution optical imagery. 
This diversity brings new possibilities for multi-modal RS applications, but it also raises the challenge of selecting the most suitable FM for a given task with data modality and operational constraints.

Despite these advances, selecting a suitable FM for a specific RS task remains a challenge. 
RS practitioners must balance diverse constraints such as the available data modalities and volume, geographic coverage, computational resources, and task-specific evaluation priorities. These constraints have been shown to significantly influence RSFM generalization and robustness~\cite{purohit2025how, Plekhanova0DABW25}. With hundreds of remote sensing foundation models~(RSFMs) now publicly available~\cite{guo2024skysense, li2025unleashing} and no unified structured schema to organize their properties~(such as model architectures, training data, or reported performance), the selection process is often manual, which is time-consuming, error-prone, and difficult to reproduce. Existing approaches rely on searching across scattered repositories and publications, manually parsing papers and model cards, and running exhaustive experiments~\cite{ramachandran2025primer,pierre2025benchmark}, all without guaranteed reproducibility or transparency. Even public RS benchmarks~\cite{geobench, geobench2, guo2024skysense} mainly compare model accuracy on fixed applications, but they do not address the upstream selection problem of identifying suitable candidate FMs under user-specific modalities, data availability, compute budgets, and deployment trade-offs. This makes a unified, machine-readable database~(DB) of RSFMs a necessary foundation for any systematic selection and automation.

\begin{figure}[t]
  \centering
   \includegraphics[width=\linewidth]{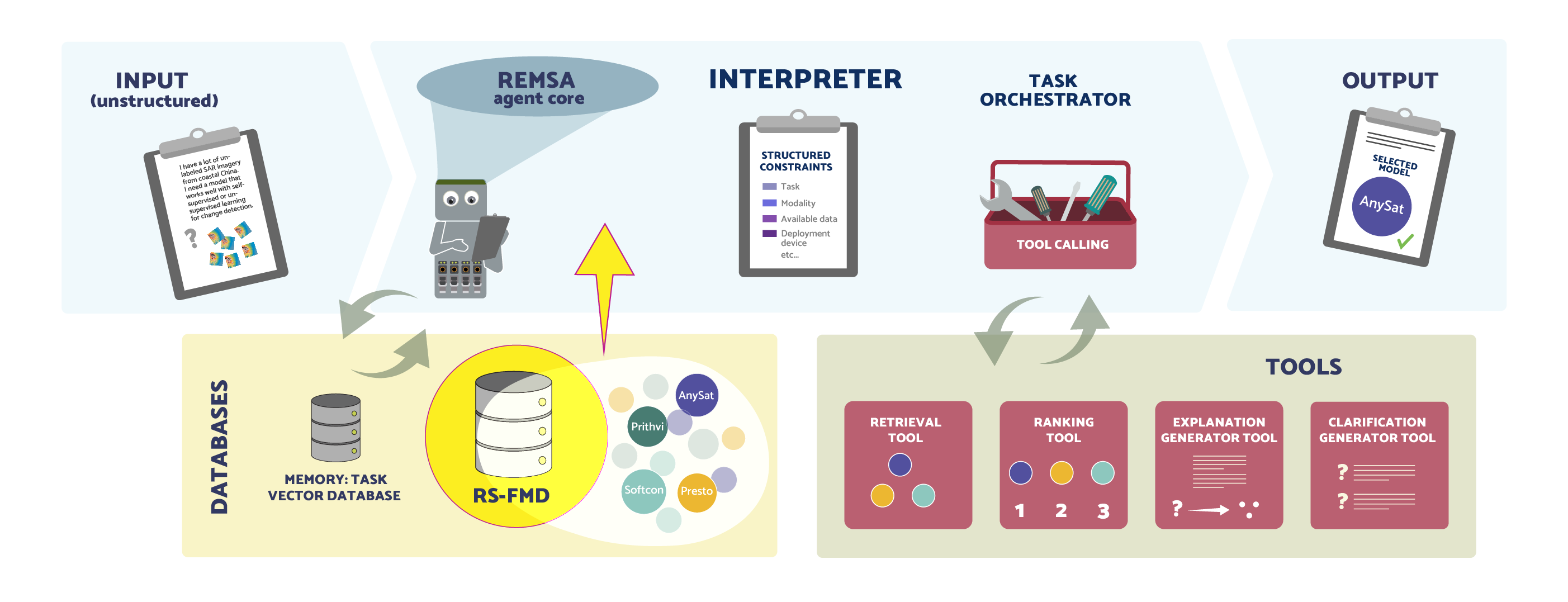}
   \caption{Architecture of \system for remote sensing foundation model selection. \system parses an unstructured user query, grounds the extracted constraints in \fmd and task memory, and orchestrates retrieval, ranking, clarification, and explanation tools to produce justified model recommendations.}
   \vspace{-19pt}
   \label{fig:framework}
\end{figure}

Recent advances in large language model~(LLM) agents have shown the potential of integrating language models with external tools and structured workflows to support complex tasks~\cite{llavaplus,omniact,videoagent1,videoagent2}. However, most existing agents target general-purpose question answering or conversational assistance. 
To the best of our knowledge, no prior work has developed a domain-specific agent for FM selection in operational, constraint-heavy RS scenarios. In particular, RS tasks involve complex trade-offs across sensors, spatial, spectral, and temporal resolutions, as well as data availability. Existing LLMs lack the domain knowledge and structured access to model documentation to address these constraints. Hence, such an agent must provide more robust and interpretable solutions.



In this work, we first introduce the RSFM Database (\fmd), the first schema-guided resource of over 160 RSFMs, covering various data modalities, pretraining strategies, and reported benchmark results. Built on \fmd, we propose \system, a first LLM-based agent for automated FM selection in RS. 
As shown in \cref{fig:framework}, \system is a modular agent that supports FM selection through structured query interpretation and task-aware tool invocation. It extracts user intent from free text input and converts it into constraints. Based on the task state, the agent orchestrates tools for retrieval from \fmd, ranking FMs, interactive clarification, and justified explanation. A lightweight memory mechanism further enhances accuracy and personalization.
To evaluate \system, we build a new benchmark of realistic user queries and establish a rubric-based expert evaluation protocol that scores the suitability of selected FMs under user-specified constraints. We also implement a set of carefully constructed baselines, ensuring fair and meaningful comparisons with \system.
\system is designed to support a broad range of users, including RS scientists, machine learning researchers, and industry practitioners who need to identify suitable RSFMs for their tasks. Because \system accepts free-text queries and incorporates structured interpretation together with multi-turn clarification, it can assist even non-experts who may not be familiar with RS modalities or FM architectures. 
This makes \system suitable for both exploratory use by practitioners and rigorous FM selection in research settings.
Although \system adopts a modular agent design rather than proposing a new visual backbone, our contribution is methodological. We formalize RSFM selection as an under-supported stage in practical RS or computer vision pipelines: deciding how FMs should be compared, selected, and deployed under modality, data, compute, and application constraints before downstream adaptation.
In summary, we make the following \textbf{contributions}:
\begin{itemize}[leftmargin=*]
    \item We introduce \fmd, the first structured and schema-guided DB of over \textbf{160} RSFMs. We will release it as a community resource with continuous maintenance and updates.
    \item We propose \system, a modular LLM agent that combines structured metadata grounding, dense retrieval, in-context ranking, clarification, explanation, memory augmentation components, and a task-aware orchestration mechanism to support complex FM selection in real RS settings.
    \item We construct a new benchmark dataset and rubric-based evaluation protocol for FM selection in RS, comprising 100 realistic query scenarios and \textbf{3,000} expert-scored evaluations across systems, LLM backbones, and selected models.
\end{itemize}


\section{Related Work}

\stitle{Foundation Models and Model Selection.}
Due to the rapid emergence of FMs, there has been extensive research into their capabilities and benchmarks~\cite{internvideo2,kerem2024do,fmorfinetune}. In RS, recent surveys and benchmarks~\cite{rsfm_survey, ramachandran2025primer, guo2024skysense} have systematically cataloged FMs and evaluated their performance on applications such as land cover classification, wildfire scar segmentation, urban change detection, visual question answering, etc. However, these works primarily focus on descriptive analysis or standardized evaluation, offering limited support for automated FM selection. 
Large-scale evaluations such as GEO-Bench-2~\cite{geobench2} further highlight that RSFM performance varies strongly across capability dimensions, but they mainly evaluate models under fixed benchmark protocols rather than addressing the upstream problem of selecting suitable FMs under user-specific constraints.
Other work also shows that pretraining data coverage~(geographic and sensor diversity) strongly affects RSFM generalization~\cite{purohit2025how, Plekhanova0DABW25}. While current benchmarks document these properties, they do not use them to guide model choice, further motivating automated FM selection.
Additionally, recent capabilities encoding approach estimates a model’s performance on unseen downstream tasks, reducing the need for exhaustive fine-tuning~\cite{pierre2025benchmark}. Although such performance prediction methods provide valuable tools for comparative evaluation, they are complementary to our setting. They rely on standardized transfer evidence and do not address end-to-end FM selection from open-ended user queries involving modalities, data availability, compute budgets, and deployment constraints.
Moreover, previous surveys and benchmarks are static and task-specific, lacking a unified schema or machine-readable representation of RSFMs. In contrast, our \fmd consolidates the available FMs into a structured, extensible resource that directly supports automated retrieval, comparison, and selection. 
Another relevant line of work is AutoML, which includes frameworks such as Auto-WEKA~\cite{autoweka}, Auto-sklearn~\cite{autosklearn}, and CAML~\cite{caml}. They automate the selection of parameters, algorithms, or pipelines through meta-learning and optimization techniques. Although these approaches show the feasibility of automating model choice in classical machine learning settings, they have not been applied to the selection of FMs, particularly in the RS domain.
To our knowledge, there is no existing dedicated method or agent that assists scientists in selecting the most suitable FM for their specific constraints and applications. Our work fills this gap by combining \fmd and \system, formalizing constraint-aware RSFM selection as an upstream step in practical RS or computer vision pipelines and implementing it through structured metadata grounding, task-aware orchestration, clarification, ranking, and explanation.

\stitle{Tool-Augmented Agents in RS.}
Recent developments in retrieval-augmented language models and tool-augmented agents such as LLaVA-Plus~\cite{llavaplus}, OmniACT~\cite{omniact}, VideoAgent~\cite{videoagent1} show the feasibility of combining LLMs with structured retrieval and external tool invocation for complex task execution and workflows. In RS, several works have explored modular agentic workflows. GeoLLM-Squad~\cite{geollmsquad} introduces a multi-agent orchestration framework that decomposes geospatial tasks into specialized sub-agents, improving scalability and correctness over single-agent baselines. RS-Agent~\cite{RS-agent} integrates retrieval pipelines and tool scheduling to process spatial question answering tasks, while ThinkGeo~\cite{thinkgeo} introduces a benchmark for evaluating multi-step tool-augmented agents on RS workflows. 
Recently, EarthDial~\cite{earthdial} extends large vision-language models to multi-sensor earth observation data through interactive dialogue modeling, enabling conversational reasoning over complex data modalities.
These agentic approaches highlight the benefits of modularity and retrieval-augmented decision process. However, they mainly target geospatial information extraction, change detection, or VQA applications rather than FM selection. 
\system explicitly integrates a curated FM database with structured retrieval, agentic ranking, interactive constraint resolution, and transparent model reasoning, making it a dedicated tool-augmented agent tailored for FM selection in RS.

\section{Remote Sensing Foundation Model Database}
\label{sec:fmd}

\fmd is a curated DB of RSFMs identified through a systematic search covering over 160 RSFMs, serving as the structured resource behind \system. It enables interpretable and constraint-aware FM selection by consolidating heterogeneous resources into a unified, machine-readable format. To build \fmd, we conducted a systematic search for RSFMs using multiple sources. We reviewed survey papers and popular FM lists, surveyed recent RS and ML venues, ran keyword searches on arXiv, and inspected linked GitHub repositories.

\stitle{Schema Design.}
Each record follows a schema covering properties such as identifiers, architecture, modalities, and pretrained model weights, along with structured fields for pretraining datasets and reported benchmark evaluations when available. This schema ensures traceability, comparability, and extensibility across FMs. The full schema and an example record are in Appendix~(App.) Sec. A.
This comprehensive schema enables our FM selection agent to ground the selection process in documented model properties and reported capabilities, and match models to user-defined tasks and constraints. It also ensures that critical properties, such as input data modalities, spatial, spectral, and temporal characteristics, and training configurations, can be queried and filtered in a principled and automated manner.

\stitle{Automated database population.}
Populating this database requires extracting structured information from diverse sources, such as papers, model cards, and repositories. Due to the scale and heterogeneity of available model documentation, fully manual curation is impractical. Therefore, we adopt a semi-automated knowledge extraction approach coupled with confidence-guided human verification. Our approach is a schema-guided LLM extraction pipeline inspired by a general knowledge extraction approach OneKE~\cite{oneke}, but significantly adapted to our domain and use case.
Specifically, we extend their approach by introducing our own schema definitions, adding a dedicated confidence scoring step, and optimizing prompt design for RS model descriptions. The process is entirely automated and iterative: for each FM, we collect and input a set of unstructured sources, then issue multiple LLM calls to generate independent structured outputs in each iteration. Each output is validated against the schema, parsed, and aggregated. This iterative strategy allows us to exploit both the probabilistic uncertainty of each iteration and the self-consistency across iterations. Fields for which the model produces divergent outputs or low log-probabilities are marked as uncertain and passed to the human verification stage.
The resulting pipeline effectively converts complex heterogeneous text sources into machine-readable JSON records with minimal targeted manual intervention.

\stitle{Confidence Score for Human Verification.}
Ensuring the reliability of the extracted metadata is critical for FM selection in downstream RS tasks. To this end, we define a confidence score for each field in each record, enabling targeted human verification only where the uncertainty is high. Our confidence score combines two complementary criteria: the model's generation probability and the consistency of outputs across multiple LLM sampling rounds.
For each field, we compute the confidence score as follows:
\begin{equation}
\begin{aligned}
\text{Confidence} =
    w_{\text{logp}} \cdot \text{NormalizedLogProb} +\, w_{\text{cons}} \cdot \text{SelfConsistency}
\end{aligned}
\end{equation}

\noindent
where \textbf{NormalizedLogProb} quantifies the LLM’s internal certainty by mapping the raw log-probability of the generated field value to a bounded range, and \textbf{SelfConsistency} measures the fraction of LLM generations that agreed on the same value among multiple independent sampling iterations.

To ensure interpretability and numerical scaling, we normalize log-probabilities using a sigmoid function controlled by the temperature parameter. We set the temperature $\tau = 0.5$ to avoid saturation and preserve sensitivity in the moderate-confidence regime of the sigmoid function. 
We set $w_{\text{logp}} = 0.7$ and $w_{\text{cons}} = 0.3$ to prioritize the log-probability signal while still leveraging the stabilizing effect of self-consistency. These weights were empirically determined via a grid search on a validation set of randomly sampled 10 FM records with manually verified ground truth. We optimized for maximum agreement between the confidence score and human verification decisions, using the area under the precision-recall curve~(PR-AUC) as the selection criterion. We observed that prioritizing the log-probability signal improved precision, while incorporating self-consistency helped identify low-confidence outliers. However, these weights are not necessarily fixed and can be adjusted by practitioners depending on the properties of their LLMs, model domains, or confidence calibration needs.
Any field with a final confidence below a threshold $\theta = 0.75$ is automatically flagged for human review. Importantly, human annotators inspect only the flagged fields rather than full model records.
Reviewing all fields for all FMs would require substantial annotation effort, as each record contains many heterogeneous metadata elements. To assess the risk of confidently incorrect extractions, we manually inspected all fields for 10 records and found high-confidence outputs to be consistently accurate, supporting the reliability of our scoring mechanism. In practice, occasional field-level errors have limited impact on FM selection, as the most decisive documented properties~(modality, architecture, compute requirements, and reported performance evidence) are usually clearly stated and rarely mis-extracted.

\stitle{Diversity of Coverage.}
The current release of \fmd spans a broad range of RSFMs pretrained on various data modalities~(multispectral, hyperspectral, SAR, LiDAR, and text) and employing diverse model architectures~(transformer-based encoders, CNN–transformer hybrids, vision–language models). Pretraining data sources range from small curated datasets to million-scale image collections, and spatial resolutions span from sub-meter imagery to coarse multispectral composites. By consolidating these heterogeneous information sources into a schema-guided resource, \fmd supports reproducible comparison, metadata-driven analysis, and automated retrieval. 
We will maintain \fmd in a public repository under a permissive license through a community-in-the-loop update process. In addition to monthly rolling checks by the maintainers for newly released RSFMs, model authors can submit documentation for new models directly. \system then automatically extracts the corresponding schema fields and presents them for inspection, while maintainers review the submitted records before inclusion to ensure consistency and reliability. This design reduces maintenance latency compared with fully manual curation, while retaining quality control over the structured metadata.

\section{\system Agent Architecture}

The goal of \system is to automate the upstream selection of FMs for RS tasks through a decision-centered, modular agentic workflow. \system integrates structured knowledge grounding, LLM-assisted ranking with in-context learning, and iterative clarification to produce transparent and reproducible selections. 
Selecting an appropriate RSFM is challenging, as the models differ in data modalities, pretraining strategies, reported benchmark evidence, and resource requirements. In addition, users often provide incomplete or ambiguous task descriptions, requiring the agent to infer user intent and reconcile trade-offs among candidate models. To address these challenges, \system provides an integrated pipeline combining different agent components and external tools. This pipeline decomposes RSFM selection into structured retrieval, constraint filtering, ranking, clarification, explanation, and memory augmentation under a customized orchestration mechanism. This section describes the agent workflow and the details of each component and tool.

\subsection{Agent Workflow}

\cref{fig:framework} illustrates the architecture of \system that is composed of two main layers: the \textbf{LLM agent core} and a set of \textbf{external tools}. The LLM agent core consists of two key components: the \emph{Interpreter}, which parses user inputs and extracts user intent, and the \emph{Task Orchestrator}, which dynamically decides which external tool to invoke at each step based on the current task state.
When a user submits a free-text query, the \emph{Interpreter} transforms it into a structured representation of constraints.
We prompt the LLM with a carefully designed schema that covers both mandatory and optional fields relevant to RSFM selection~(see App.~Sec.~B for complete schema).
Specifically, the parser extracts the target \texttt{application}~(\eg, land cover classification, surface water segmentation) and the required \texttt{modality}~(\eg, multispectral, SAR) as mandatory fields to narrow the FM search space.
Then \system integrates broader practical constraints through optional fields and clarification steps, including data availability, compute budget, fine-tuning requirements, and output quality priorities.
Once constraints are available, the \emph{Task Orchestrator} initiates a control loop that manages the entire selection process. At each step, it first evaluates the current task state, including constraint completeness, candidate set size, hard constraint violations, and ranking confidence. Then it invokes the appropriate tool accordingly. If no mandatory constraints are missing, the orchestrator calls the \emph{Retrieval Tool} to generate an initial candidate set. If the candidate set is small and all constraints are satisfied, the \emph{Ranking Tool} is applied directly. If there are too many candidates or if ranking results yield low confidence scores, the orchestrator calls the \emph{Clarification Generator Tool} to ask the user for additional input. The updated query is then passed back through the same loop. Once the top-$k$ result is obtained, the \emph{Explanation Generator Tool} is invoked to produce the final report.
This decision process is governed by predefined thresholds for ranking confidence, constraint coverage, candidate set size, and clarification rounds. The orchestration makes the selection process adaptive and transparent, rather than a single-step retrieval or unconstrained LLM recommendation.
To support personalization and long-term refinement, \system also integrates \emph{Task Memory}, which stores past user interactions in a lightweight vector database. Relevant memory entries are retrieved via cosine similarity to improve future interactions.
More details on the implemented workflow are in App.~Sec.~C.
To enhance \system's reliability, we have several built-in mechanisms to mitigate failures. The orchestrator monitors confidence signals and triggers clarification rounds when ranking is uncertain. Rule-based constraints filter candidates that violate hard requirements. A fallback ``closest-matc'' mode returns the most compatible alternative when no candidate fully satisfies the constraints. Our modular design also allows for integrating explicit feedback mechanisms~(\eg, an optional LLM-as-a-Judge component that re-evaluates low-quality selections), making \system extensible to more robust error mitigation strategies.

\subsection{Agent Tools}
\label{subsec:tools}
The following tools operate as callable interfaces outside of the agent core. Each tool is invoked independently by the orchestrator, depending on the task state, supporting retrieval, ranking, clarification, and explanation within the RSFM selection workflow.
Our design supports extensibility for future tool integration.

\stitle{Retrieval Tool.}
To generate an initial candidate set, the retrieval tool encodes both the structured constraints and the FM entries in the \fmd using Sentence-BERT embeddings~\cite{sentencebert}. To preserve the metadata structure in the embedding, each metadata field is prefixed with a type-indicator token~(\eg, [APPLICATION], [MODALITY]) before encoding.
\system uses Facebook AI Similarity Search~(FAISS)~\cite{faiss} for an efficient search based on cosine similarity. The tool returns a list of the most relevant FMs determined by a configurable similarity threshold. Users can adjust it based on their domain requirements. In our experiments, we set this threshold empirically based on preliminary experiments to ensure broad coverage while minimizing irrelevant matches. This tool is optimized for high recall: it includes soft matches and does not enforce strict constraints, allowing the downstream pipeline to handle finer filtering.

\stitle{Ranking Tool.}
While the retrieval tool provides a broad list of relevant FMs, it cannot fully capture user-specific needs and deployment trade-offs. The ranking tool therefore combines deterministic constraint filtering with LLM-based preference ordering over the remaining candidates. The ranking tool refines the candidate FM list using a hybrid strategy to balance efficiency, flexibility, and interpretability:

\begin{itemize}[leftmargin=*]
    \item \emph{Rule-Based Filtering:} Candidates that violate hard constraints, such as required modality, sensor support, or mandatory deployment requirements, are eliminated using deterministic logic. These hard constraints are defined based on fields extracted by the interpreter.
    
    \item \emph{In-Context LLM Ranking:} The remaining candidates are re-ranked by an LLM prompted with the structured query and FM metadata, using expert-crafted few-shot examples that demonstrate how to balance application fit, modality compatibility, reported evidence, and deployment constraints. The LLM returns an ordered list with brief justifications, leveraging in-context ranking rationale without any model training (details in App.~Sec.~D). We also compute a field-aware confidence score for each selection following \cref{sec:fmd}.
\end{itemize}

\stitle{Clarification Generator Tool.}
If the orchestrator detects insufficient constraints or a low overall confidence score of selected FMs, it invokes the clarification tool. This tool inspects the parsed schema to determine missing or under-specified fields~(\eg, modality, region, compute budget, or output-quality priority) and formulates clarification questions. The tool generates each question based on the interpreter schema. We limit the clarification to three rounds to avoid user fatigue. \system will integrate the responses with initial user input, parse and merge them into the evolving task specification, in order to iteratively refine the selection process.

\stitle{Explanation Generator Tool.}
Once a ranking is available, this tool generates structured, human-readable explanations. It uses a prompt-driven LLM to synthesize the rationale for each selected FM, including key reasons about suitability, constraint satisfaction, and trade-offs among candidates. Each JSON format output includes the model name, bullet points for explanation, and links to the corresponding paper and code repository. This tool enhances transparency and user trust by exposing the selection process~(prompt is in App.~Sec.~E). 

\section{Evaluation Protocol and Benchmark}
\label{sec:evaluation}

Evaluating FM selection in RS is challenging due to the lack of dedicated benchmarks. 
Previous works mainly focus on evaluating model accuracy after adaptation on fixed tasks or datasets, rather than assessing the upstream ability to recommend suitable FM under diverse real-world deployment constraints. 
In this work, we leverage \fmd to construct an agent-based benchmark for upstream FM selection, systematically covering diverse models, modalities, and deployment constraints.

\stitle{Benchmark Construction.}
Our evaluation protocol relies on structured expert scoring rather than agreement with human-selected answers, ensuring methodological rigor without imposing excessive annotation overhead.
We curate a benchmark of 100 diverse natural language queries to keep expert evaluation feasible while still ensuring diversity and realism. All queries can be found in our source code.
Each selected model-query pair is anonymized with respect to the generating system and evaluated by two experts with complementary RS and ML expertise. 
Experts do not provide a separate ground-truth top-1 FM. Instead, they score the suitability of each selected model under the user-specified constraints using a structured rubric.
Full details of the expert procedure are in App.~Sec.~G.
In total, the evaluation yields 3,000 expert ratings, as experts assess the top-3 FMs selected by \system and 3 baseline systems across the query set. 
Each instance is carefully rated across seven criteria, resulting in a substantial expert evaluation workload despite the modest number of unique queries.
To maximize representativeness, we create the queries using structured templates of various scenarios and instantiate them~(Full templates are in App.~Sec.~H.) The queries diversify over data availability, computational resources, application complexity, and evaluation priorities. 
The resulting queries cover a wide range of tasks, including flood mapping with SAR data, crop type classification using multispectral or hyperspectral imagery, urban expansion monitoring with optical time series, and environmental monitoring or hazard-related applications, such as sea ice and wildfire detection. These tasks cover both single-date and multi-temporal analysis, single- and multi-modal inputs, and varying resource environments.
All queries were reviewed by a domain expert for factual accuracy and corrected for consistency.

\stitle{Baselines.}
There is limited prior work on automated FM selection for RS deployment, and existing AutoML or agent systems cannot directly perform this task. We therefore design baselines that serve as both meaningful comparisons and component analysis of \system, with each baseline removing or modifying specific components to assess their contributions:


\begin{enumerate}[leftmargin=*]
    \item \underline{\textsc{\system-Naive}}: Same toolset and DB as \system, but employs only basic sequential orchestration without \system's adaptive, task-aware control logic. It relies on LangChain's default single-step execution, where the LLM independently chooses tools without structured workflow or multi-turn coordination~\cite{langchain}. This baseline tests the effectiveness of our orchestration mechanism. 
    \item \underline{\textsc{DB-Retrieval}}: Returns the top-$k$ models from the FAISS-based dense retrieval over \fmd, with ranking, clarification, memory, and orchestration removed. This serves as a retrieval-only baseline and isolates the contribution of LLM-based ranking and constraint reasoning.
    \item \underline{\textsc{Unstructured-RAG}}: A generic RAG setup where LLM receives the query and unstructured FM descriptions and outputs top-$k$ FMs with brief justifications (prompt in App.~Sec.~F). This baseline tests whether a general-purpose LLM can perform FM selection without our structured, modular agent.
\end{enumerate}

\begin{table}[t]
\caption{Expert rubric for scoring the suitability of selected query--model pairs.}
\label{tbl:criteria}
  \centering
    \begin{tabular}{ll}
    \toprule
    \textbf{Criterion} & \textbf{Description} \\
    \midrule
    Application Compatibility & Whether the model fits the user requested application \\
    Modality Match & Whether the model supports the required input data modality \\ 
    Reported Performance & Performance reported on similar datasets or applications \\
    Efficiency & Suitability for the user's computational resources \\
    Popularity & Based on GitHub repository stars and citations \\
    Generalizability & Diversity and scale of pretraining data \\
    Recency & Whether the model reflects recent developments \\
    \bottomrule
    \end{tabular}
    \vspace{-8pt}
\end{table}

\stitle{Evaluation Protocol and Criteria.}
For each query, \system and all baselines output their top-3 selections. These model-query pairs were evaluated independently and blindly by the two experts using the 7 criteria in \cref{tbl:criteria}. 
After independent scoring, substantial disagreements were resolved according to the predefined rubric, with the corresponding model documentation re-examined when necessary. 
The evaluation protocol was fixed before scoring, and no candidate sets, system outputs, rubric criteria, or model records were modified after scoring to avoid post-hoc bias.
Each selected FM was rated on a 1-5 scale~(0.5 precision) across the 7 criteria, covering task relevance and deployment suitability under real-world constraints. 
Several criteria use explicit rules. For example, generalizability combines geographic, modality, and dataset-scale factors, popularity relies on citations or GitHub activity, and recency is based on publication year~(More details are in App.~Sec.~G.).
The criteria are designed to be transparent, reproducible, and grounded in practical needs rather than ad-hoc user preferences. 
The final score is a weighted sum of all criteria ratings~(Our weight settings are in App.~Sec.~I.). The score is linearly mapped to a 1-100 scale to better show the differences. 

Exhaustive empirical benchmarking of all candidate models under all query-specific settings is infeasible, because such downstream accuracy--cost benchmarking would require adapting and evaluating many heterogeneous RSFMs for each user-specific task, modality, region, and resource setting, which is beyond the scope of this work. Instead, we develop a protocol that offers a reproducible and practical proxy for evaluating the quality of real-world FM selection workflows.
To support broader community adoption, we publicly release the full set of evaluation queries, expert guidelines, scoring criteria, and model metadata used in the evaluation. This enables reproducibility and provides a standardized foundation for future research on FM selection in RS and beyond.
Our evaluation does not assume a single ground-truth ``best'' FM. Instead, experts score each selected model-query pair according to a fixed rubric, and systems are compared based on the suitability scores of their returned candidates. Thus, the evaluation measures model selection quality under user constraints, rather than agreement with independently selected human answers or downstream task accuracy after fine-tuning. \system returns top-$k$ FMs with explanations, enabling users to choose based on their own preferences.

\section{Experiments}
\label{sec:experiments}

We conduct experiments to comprehensively evaluate the quality of \system's RSFM recommendations under user-specified constraints. Since limited prior work directly targets real FM selection under diverse deployment constraints, we develop our own baselines. This section presents our experiment setup, quantitative results, and sensitivity analysis, followed by a discussion of limitations and representative examples.

\stitle{Experimental Setup.}
We use \texttt{GPT-4.1}~\cite{gpt4.1} as the primary LLM core for \system and all baselines to ensure fairness. 
To evaluate robustness across different backbones, we additionally conduct experiments with \texttt{DeepSeek3.2}~\cite{deepseek} and \texttt{LLaMA-3.3-70B}~\cite{llama}. 
\system is designed to be LLM-agnostic and can readily support different LLMs without architectural changes. Our benchmark consists of 100 diverse natural language user queries. For each input, \system and all baselines~(all described in \cref{sec:evaluation}) select the top-$3$ candidate FMs for comparison. Domain experts rate each candidate using the criteria in~\cref{tbl:criteria}, and we report both single-model and set-level suitability scores to evaluate overall selection quality.
During evaluation, all clarification rounds in \system were executed automatically, with the system interacting with a separate LLM simulating user responses. No human was involved in these interactions, ensuring consistency and preventing evaluator bias across systems and LLM backbones.

\stitle{Evaluation Metrics.}
We use complementary metrics to evaluate both the top-ranked model and the overall returned set quality:
(1) \emph{Average Top-1 Score}~(expert-assigned suitability score of the top-ranked model),  
(2) \emph{Average Set Score}~(average expert-assigned suitability score of the top-3 models),  
(3) \emph{Top-1 Hit Rate}~(fraction where the system's top-ranked model receives the highest expert score among pooled candidates),  
(4) \emph{High-Quality Hit Rate}~(fraction where the top model scores $\geq 80$), and  
(5) \emph{Mean Reciprocal Rank - MRR}~(rank of the highest-scored pooled candidate within the system's top-3).

\begin{table}[t]
\centering
  \caption{Model selection quality comparison to the baselines~(GPT-4.1)}
  \label{tbl:baseline_results(GPT-4.1)}
    \begin{tabular}{lccccc}
    \toprule
    \textbf{System} & \textbf{Avg Top-1} & \textbf{Avg Set} & \textbf{Top-1 Hit} & \textbf{HQ Hit} & \textbf{MRR} \\
    \midrule
    \system (Ours) & \textbf{75.76} & \textbf{75.03}  &\textbf{21.33}\% & \textbf{40.00\%} & \textbf{0.34} \\
    \textsc{\system-Naive} &72.67  &72.00   & 20.00\%  &37.33\% &0.29  \\
    \textsc{DB-Retrieval}  &67.37   &68.87  &12.00\%   &17.33\% &0.23\\
    \textsc{Unstr.-RAG} &71.23   &68.39  &13.33\%  &30.67\% &0.24\\
    \bottomrule
    \end{tabular}
\vspace{-10pt}
\end{table}

\subsection{Comparison to baselines}

As shown in \cref{tbl:baseline_results(GPT-4.1)}, \system consistently outperforms all baselines across the evaluated metrics, demonstrating its effectiveness for FM selection under various real constraints. 
Illustrative examples of expert-scored model-query pairs are provided in App.~Sec.~J.
Under GPT-4.1, \system achieves the highest Average Top-1 Score~(75.76) and Average Set Score~(75.03), indicating not only that the top-ranked models receive stronger suitability scores, but also that the top-3 selections offer competitive and diverse alternatives. 
It also achieves the highest Top-1 Hit Rate~(21.33\%), High-Quality (HQ) Hit Rate~(40.00\%), and MRR~(0.34), indicating both precision at rank 1 and stable ranking quality.
Compared to \textsc{DB-Retrieval}, which relies on similarity-based retrieval over structured metadata, \system improves Top-1 Hit Rate from 12.00\% to 21.33\%, HQ Hit Rate from 17.33\% to 40.00\%, and MRR from 0.23 to 0.34. This underscores the value of structured decision logic beyond retrieval, especially when user queries involve constraints~(\eg, modality, resolution, compute budget) that require the composition of multiple constraints rather than direct metadata matching.
Although \textsc{Unstr.-RAG} has access to full model descriptions, its performance remains lower due to the lack of structured guidance and modular reasoning. While it achieves moderate Average Top-1 scores~(71.23) compared to \system, its Top-1 Hit Rate~(13.33\%) and MRR~(0.24) remain substantially below \system. This result shows that \system's ability to combine structured schema grounding with task-aware tool orchestration enables more precise alignment with user needs.
Both \system and \textsc{\system-Naive} perform notably better than retrieval-only or unstructured RAG baselines, showing the effectiveness of our modular architecture: grounding the selection process in a structured schema and enabling tool-based reasoning provides a substantial advantage.
However, \system improves further across all major evaluation metrics compared to \textsc{\system-Naive}. For example, Average Top-1 Score improves from 72.67 to 75.76, HQ Hit Rate from 37.33\% to 40.00\%, and MRR from 0.29 to 0.34. These gains suggest that our orchestration logic, including multi-turn clarification and decision heuristics, contributes meaningfully to selection quality, especially when model choices are ambiguous or task formulations are complex.

\stitle{Latency Trade-off.} To assess the latency-performance trade-off, we measure the average end-to-end runtime per query. As expected, single-step methods are faster: DB-Retrieval takes 0.77s, Unstr.-RAG 11.9s, and \system-Naive 22.7s, whereas \system requires $\text{\textbf{31.7s}}$ due to multi-stage decision processing and optional clarification. Despite this moderate overhead, \system delivers the highest expert-scored selection quality across major metrics, indicating that its additional reasoning steps yield meaningful and consistent gains.

\subsection{Effect of different LLM backbones}

\begin{table}[t]
\centering
  \caption{Model selection quality comparison to the baselines~(DeepSeek3.2)}
  \label{tbl:baseline_results(DeepSeek3.2)}
    \begin{tabular}{lccccc}
    \toprule
    \textbf{System} & \textbf{Avg Top-1} & \textbf{Avg Set} & \textbf{Top-1 Hit} & \textbf{HQ Hit} & \textbf{MRR} \\
    \midrule
    \system (Ours) & \textbf{75.35} & \textbf{73.81} & \textbf{18.67\%} & \textbf{40.00\%} & \textbf{0.30} \\
    \textsc{\system-Naive} &72.03  &71.83   &16.51\%  &36.89\% &0.26  \\
    \textsc{DB-Retrieval}  &67.37   &68.87  &12.00\%   &17.33\% &0.23\\
    \textsc{Unstr.-RAG} &69.19   &70.94  &10.67\%   &24.00\% &0.24\\
    \bottomrule
    \end{tabular}
\end{table}

\begin{table}[t]
\centering
  \caption{Model selection quality comparison to the baselines~(LLaMA-3.3-70B)}
  \label{tbl:baseline_results(LLaMA-3.3-70B)}
    \begin{tabular}{lccccc}
    \toprule
    \textbf{System} & \textbf{Avg Top-1} & \textbf{Avg Set} & \textbf{Top-1 Hit} & \textbf{HQ Hit} & \textbf{MRR} \\
    \midrule
    \system (Ours) & \textbf{73.39} & \textbf{70.34} & \textbf{14.67\%} & \textbf{32.00\%} & \textbf{0.26} \\
    \textsc{\system-Naive} &69.02  &69.00   &14.23\%  &29.47\% &0.24 \\
    \textsc{DB-Retrieval}  &67.37   &68.87  &12.00\%   &17.33\% &0.23\\
    \textsc{Unstr.-RAG} &69.87   &68.04  &10.00\%  &26.67\% &0.22\\
    \bottomrule
    \end{tabular}
\vspace{-10pt}
\end{table}

To evaluate the robustness of \system under different LLM backbones, we further compare results using GPT-4.1, DeepSeek3.2, and LLaMA3.3-70B. As shown in \cref{tbl:baseline_results(GPT-4.1)}, \cref{tbl:baseline_results(DeepSeek3.2)}, and \cref{tbl:baseline_results(LLaMA-3.3-70B)}, across all three cores, \system consistently outperforms the corresponding baselines, indicating that the improvements primarily stem from the overall architecture rather than a specific LLM. 
Selection quality is strongest with GPT-4.1~(Avg Top-1: 75.76; MRR: 0.34), followed closely by DeepSeek3.2~(75.35; 0.30), while LLaMA3.3-70B shows slightly lower but still competitive results~(73.39; 0.26). Notably, HQ Hit Rate remains relatively stable across GPT-4.1 and DeepSeek3.2 (both 40.00\%), suggesting that high-quality candidate identification is robust to backbone choice. The consistent margin over \textsc{\system-Naive} and retrieval-based baselines across all LLM backbones demonstrates that structured grounding and tool-based orchestration provide consistent gains independent of the underlying language model.

\subsection{Sensitivity Analysis on Evaluation Criteria}

\begin{table}[t]
\centering
  \caption{Sensitivity of expert-scored suitability metrics to individual rubric criteria.}
  \label{tbl:sensitivity_results}
    \begin{tabular}{lcccc}
    \toprule
    \textbf{Criteria Setting} & \textbf{Avg Set} & \textbf{Top-1 Hit} & \textbf{MRR} & \textbf{Note} \\
    \midrule
    Full Scoring (All Criteria) &75.03  &22.67\% &0.38 & \multirow{7}{*}{\parbox{1cm}{\textcolor{goodgreen}{Green:} \\Increase \\\textcolor{badred}{Red:} \\Drop}} \\
    w/o Application Compatibility &\textcolor{badred}{73.32}   &\textcolor{badred}{21.33\%}  &\textcolor{badred}{0.36}   &   \\
    w/o Modality Match &\textcolor{badred}{70.88}   &22.67\%  &\textcolor{badred}{0.36}   &     \\
    w/o Reported Performance &\textcolor{goodgreen}{75.05}   &22.67\%  &0.38   &     \\
    w/o Efficiency &\textcolor{goodgreen}{80.23}   &\textcolor{goodgreen}{25.33\%}  &0.38   &  \\
    w/o Popularity+Recency &\textcolor{goodgreen}{75.13}   &\textcolor{goodgreen}{25.33\%}  &\textcolor{goodgreen}{0.39}   &    \\
    w/o Generalizability &\textcolor{goodgreen}{75.10}   &22.67\%  &0.38   &   \\
    \bottomrule
    \end{tabular}
\vspace{-10pt}
\end{table}

To understand which rubric dimensions drive the observed selection quality, we perform a sensitivity analysis by removing each scoring criterion individually from the expert evaluation protocol. As shown in \cref{tbl:sensitivity_results}, the performance is generally robust in most dimensions, but some removals reveal important insights into which criteria contribute the most to effective model selection.
Both the removal of Application Compatibility and Modality Match lead to notable performance drops, confirming that \system actively prioritizes functionally appropriate models aligned with the user’s objective.
Notably, removing Reported Performance and Generalizability yields minimal change in overall results, implying that these dimensions are either captured implicitly through other criteria or are less decisive in the current benchmark setup. In contrast, removing Efficiency or Popularity+Recency actually leads to a modest performance gain. 
This suggests that while these criteria add practical relevance for deployment, they may occasionally favor well-known or resource-efficient models over candidates that better satisfy the core application and modality constraints.
The sensitivity results further suggest that \system does not primarily rely on superficial indicators such as citations or recency, but instead emphasizes core compatibility and constraint satisfaction in its final decisions.

\section{Conclusion and Discussion}

We proposed \system, an LLM agent that combines a structured FM database with task-aware orchestration for real RSFM selection problems.
By orchestrating modular tools for retrieval, ranking, clarification, explanation, and memory augmentation, \system enables transparent and consistent RSFM selections under user-specified constraints. 
Our structured RSFM database \fmd consolidates heterogeneous descriptions for transparent retrieval and comparison. 
Across our rubric-based expert benchmark, \system outperforms retrieval-only, unstructured RAG, and naive agent baselines in selection quality. 
This work opens several directions for future research, including expanding the benchmark to more complex scenarios, developing unified execution interfaces and lightweight accuracy--cost benchmarks for well-standardized RSFM categories, incorporating learned performance-prediction modules when sufficient standardized transfer data becomes available, and exploring adaptive FM selection strategies.

Although \system performs well in RSFM selection, several \emph{limitations} remain. 
Our benchmark includes 100 expert-scored queries, which may miss rare or emerging use cases, although the evaluation requires substantial effort with 3,000 expert ratings. 
In addition, the ranking relies on in-context learning rather than supervised training, which may limit selection quality on complex queries. 

\section*{Acknowledgements}
This work is supported by the European Research Council Project Agent-BigEarth~(No. 101292498) and the European Union Horizon Project PROTEUS~(No. 101296397).

%
%
\bibliographystyle{splncs04}
\bibliography{main}

\newcommand{\papertitle}{\system: Foundation Model Selection for Remote Sensing via a Constraint-Aware Agent}
\makeatletter
\newcommand{\supplementarytitle}{%
  \begin{center}%
    {\Large \bfseries\boldmath \pretolerance=10000 \papertitle\par}\vskip .3cm
    {\large \bfseries\boldmath Supplementary Material \par}\vskip .8cm
  \end{center}%
}
\makeatother

\clearpage
\supplementarytitle

\noindent\textbf{Appendix Overview}

\begin{itemize}
    \item \cref{appendix:schema}: Complete \fmd Schema Specification.
    \item \cref{appendix:query_schema}: Structured Query Schema.
    \item \cref{appendix:implementation}: Implementation Details.
    \item \cref{appendix:icl_prompt}: LLM-Based In-Context Ranking Prompt.
    \item \cref{appendix:explanation_prompt}: Explanation Generator Prompt.
    \item \cref{appendix:rag_prompt}: Prompt for RAG-LLM Baseline.
    \item \cref{appendix:expert_evaluation}: Expert Evaluation Procedure.
    \item \cref{appendix:templates}: Query Template for Creating Benchmark Dataset.
    \item \cref{appendix:weights}: Expert Scoring Weight Configuration.
    \item \cref{appendix:expert_examples}: Illustrative Examples of Expert Scoring.
\end{itemize}

\vspace{1em}

\section*{Appendix}
\setcounter{section}{0}
\renewcommand{\thesection}{\Alph{section}}

\section{Complete \fmd Schema Specification}
\label{appendix:schema}

To properly represent the properties of each FM, we designed a comprehensive data schema for \fmd. 
The schema includes the essential characteristics of model architectures, pretraining strategies, supported modalities, and benchmark performance.

Each model record includes fields such as unique identifiers, names, versions, release and update timestamps, and links to associated publications, code repositories, and pretrained weights. These metadata elements ensure traceability and reproducibility of the database entries.

Beyond these core descriptors, the schema incorporates detailed fields that capture architectural specifics~(\eg, backbone type, number of layers, number of parameters), pretraining approaches~(\eg, pre-text training type, masking strategy), and modality integration. The design anticipates the diversity of RS models and supports future extensions.

To capture information about pretraining and evaluation comprehensively, the schema defines two nested structures:

\begin{itemize}[leftmargin=*]
    \item \textbf{PretrainingPhase}: This substructure records the datasets used for pretraining, geographical coverage, time range, image resolutions, token sizes, augmentation strategies, sampling methods, and masking ratios.
    \item \textbf{Benchmark}: This substructure captures evaluation metrics, including the applications, dataset, performance scores, and training hyperparameters used during evaluation.
\end{itemize}

Many fields are annotated with \texttt{free\_text} metadata. This annotation signals that the field may contain natural language summarization that requires specialized treatment in confidence scoring and downstream verification.

\cref{tbl:full_schema} provides a comprehensive description of the fields of our data schema in \fmd, including nested structures for pretraining phases and benchmarks.

{\scriptsize
\begin{longtable}{p{3.5cm} p{3cm} p{5cm}}
\caption{Complete schema specification of \fmd, including nested pretraining phases and benchmarks.}
\label{tbl:full_schema}\\
\toprule
\textbf{Field} & \textbf{Type} & \textbf{Description} \\
\midrule
\endfirsthead
\toprule
\textbf{Field} & \textbf{Type} & \textbf{Description} \\
\midrule
\endhead

\bottomrule
\endlastfoot

\multicolumn{3}{c}{\emph{Main Model Fields}} \\
\midrule
model\_id & string & Unique identifier of the model (free text) \\
model\_name & string & Only the name of the model without extra descriptions (free text) \\
version & string & Version identifier (free text) \\
release\_date & date & Release date of the model \\
last\_updated & date & Last updated date \\
short\_description & string & Short summary describing the model (free text) \\
paper\_link & URL & URL to the associated publication \\
citations & integer & Number of citations \\
repository & URL & URL to the code repository \\
weights & URL & URL to pretrained model weights \\
backbone & string & Specific backbone used (free text) \\
num\_layers & integer & Number of layers \\
num\_parameters & float & Model size in millions of parameters \\
pretext\_training\_type & string & Type of pretext training strategy (free text) \\
masking\_strategy & string & Masking strategy applied during training (free text) \\
pretraining & string & Description of pretraining approach (free text) \\
domain\_knowledge & list[string] & Domain-specific knowledge or methods incorporated \\
backbone\_modifications & list[string] & Modifications made to the backbone \\
supported\_sensors & list[string] & Supported satellite sensors \\
modality\_integration\_type & string & Integration type (free text) \\
modalities & list[string] & Input data modalities (free text) \\
spectral\_alignment & \{full, partial, none\} & Whether the model models spectral continuity \\
temporal\_alignment & \{full, partial, none\} & Whether the model models temporal sequences \\
spatial\_resolution & string & Spatial resolution of data (free text) \\
temporal\_resolution & string & Temporal resolution of data (free text) \\
bands & list[string] & Spectral bands used \\
\midrule
\multicolumn{3}{c}{\emph{Nested: PretrainingPhase}} \\
\midrule
dataset & string & Dataset used for pretraining (free text) \\
regions\_coverage & list[string] & Geographical regions covered \\
time\_range & string & Time range of pretraining data (free text) \\
num\_images & integer & Number of images used \\
token\_size & string & Token size (free text) \\
image\_resolution & string & Input image resolution (free text) \\
epochs & integer & Number of epochs \\
batch\_size & integer & Batch size \\
learning\_rate & string & Learning rate (free text) \\
augmentations & list[string] & Augmentations applied \\
processing & list[string] & Additional preprocessing steps \\
sampling & string & Sampling strategy (free text) \\
processing\_level & string & Processing level (free text) \\
cloud\_cover & string & Cloud cover filtering (free text) \\
missing\_data & string & Handling of missing data (free text) \\
masking\_ratio & float & Masking ratio \\
\midrule
\multicolumn{3}{c}{\emph{Nested: Benchmark}} \\
\midrule
application\_type & string & Type of application evaluated (free text) \\
application & string & Specific application domain (free text) \\
dataset & string & Benchmark dataset name (free text) \\
metrics & list[string] & List of evaluation metrics \\
metrics\_value & list[float] & Numeric values for each metric \\
sensor & list[string] & Sensors used \\
regions & list[string] & Regions evaluated \\
original\_samples & integer & Total number of samples before sampling \\
num\_samples & integer & Actual number of samples used \\
sampling\_percentage & float & Fraction of dataset retained (0–100) \\
num\_classes & integer & Number of classes \\
classes & list[string] & Names of each class \\
image\_resolution & string & Input image resolution (free text) \\
spatial\_resolution & string & Spatial resolution (free text) \\
bands\_used & list[string] & Bands used during evaluation \\
augmentations & list[string] & Data augmentations applied \\
optimizer & string & Optimizer used (free text) \\
batch\_size & integer & Batch size \\
learning\_rate & float & Learning rate \\
epochs & integer & Number of epochs \\
loss\_function & string & Loss function (free text) \\
split\_ratio & string & Train/val/test split ratio (free text) \\
\bottomrule
\end{longtable}
}

Below we include a complete example of an \fmd record for the RSFM \textit{A2-MAE}. This illustrates how the schema is instantiated with real metadata.

\begin{lstlisting}[language=json,basicstyle=\ttfamily\small,breaklines=true]
{
  "model_id": "A2-MAE",
  "model_name": "A2-MAE",
  "version": "v1",
  "release_date": "2024-06-16",
  "last_updated": "2024-06-16",
  "short_description": "A2-MAE is a spatial-temporal-spectral unified remote sensing pre-training method based on an anchor-aware masked autoencoder. It leverages a global-scale, multi-source dataset (STSSD) and introduces an anchor-aware masking strategy and a geographic encoding module to efficiently integrate spatial, temporal, and spectral information from diverse remote sensing imagery.",
  "paper_link": "https://arxiv.org/abs/2406.08079",
  "citations": 7,
  "backbone": "ViT-Large",
  "pretext_training_type": "Masked Autoencoder (MAE) with anchor-aware masking and geographic encoding",
  "masking_strategy": "Anchor-aware masking (AAM): dynamically adapts masking...",
  "pretraining": "Self-supervised pre-training on the STSSD dataset...",
  "domain_knowledge": [
    "Geographic encoding (latitude, longitude, GSD)",
    "Spatial-temporal-spectral relationships",
    "Clustering-based data pruning"
  ],
  "supported_sensors": [
    "Sentinel-2", "Landsat-8", "Gaofen-1", "Gaofen-2"
  ],
  "modality_integration_type": "Homogeneous Multimodal",
  "modalities": ["Multispectral", "Multi-temporal"],
  "spectral_alignment": "partial",
  "temporal_alignment": "partial",
  "spatial_resolution": "0.8-30m",
  "temporal_resolution": "2020-2023, periodic seasonal revisits",
  "bands": [
    "Sentinel-2: B1-B12",
    "Landsat-8: B1-B7",
    "Gaofen-1: B1-B4",
    "Gaofen-2: B1-B4"
  ],
  "pretraining_phases": [
    {
      "dataset": "STSSD",
      "regions_coverage": ["Global (12k urban centers, 10k nature reserves)"],
      "time_range": "2020-2023",
      "num_images": 2500000,
      "token_size": "16x16",
      "image_resolution": "0.8-30m (cropped 256x256 to 3200x3200)",
      "epochs": 130,
      "batch_size": 1024,
      "learning_rate": "1e-4 (cosine decay)",
      "processing": [
        "Atmospheric/radiation correction",
        "Pan-sharpening (Gaofen)",
        "Cropping/resizing alignment"
      ],
      "sampling": "Clustering-based pruning (keep hardest 10%)",
      "cloud_cover": ">=10%",
      "masking_ratio": 0.75
    }
  ],
  "benchmarks": [
    {
      "task": "Classification",
      "application": "Land cover classification",
      "dataset": "EuroSAT",
      "metrics": ["Accuracy"],
      "metrics_value": [99.09],
      "sensor": ["Sentinel-2"],
      "regions": ["34 European countries"]
    },
    {
      "task": "Classification",
      "application": "Multi-label classification",
      "dataset": "BigEarthNet",
      "metrics": ["mAP"],
      "metrics_value": [83.0]
    },
    {
      "task": "Segmentation",
      "application": "Surface water segmentation",
      "dataset": "Sen1Floods11",
      "metrics": ["mIoU"],
      "metrics_value": [88.87]
    },
    {
      "task": "Segmentation",
      "application": "Cropland segmentation",
      "dataset": "CropSeg",
      "metrics": ["mIoU"],
      "metrics_value": [44.81]
    },
    {
      "task": "Change Detection",
      "application": "LEVIR-CD",
      "dataset": "LEVIR-CD",
      "metrics": ["mIoU"],
      "metrics_value": [84.32]
    },
    {
      "task": "Change Detection",
      "application": "Urban change detection",
      "dataset": "OSCD",
      "metrics": ["F1"],
      "metrics_value": [53.97]
    },
    {
      "task": "Change Detection",
      "application": "Semantic change segmentation",
      "dataset": "DynamicEarthNet",
      "metrics": ["mIoU"],
      "metrics_value": [46.0]
    }
  ]
}
\end{lstlisting}


\section{Structured Query Schema}
\label{appendix:query_schema}

Below we show the complete JSON schema template used by the query interpreter:

\begin{verbatim}
{
  "application": "string",                          // Mandatory
  "modality": "string",                      // Mandatory
  "sensor": "string or list of strings",     // Optional
  "spatial_resolution": "string or numeric", // Optional
  "temporal_resolution": "string or numeric",// Optional
  "bands": "list of strings",                // Optional
  "avaliable_data": "string",                // Optional
  "deployment_device": "string",             // Optional
  "priority_metrics": "list of string",      // Optional
  "min_performance": {                       // Optional
    "metric": "list of string",
    "value": "list of number"
  },
  "region": "string or list of strings",     // Optional
  "domain_keywords": "list of strings"       // Optional
}
\end{verbatim}

\section{Implementation Details}
\label{appendix:implementation}

\begin{algorithm}[H]
\scriptsize
\caption{\system Workflow for RSFM Selection}
\label{al: workflow}
\KwIn{User Query $q$, desired number of recommendations $k$}
\KwOut{Top-$k$ selected models with explanations}

Initialize $ClarifyCounter \gets 0$ \\
Initialize $MaxClarify \gets 3$ \\

\Repeat{\textnormal{All mandatory constraints are present}}{
    $Constraints \gets$ \textbf{ParseQuery}$(q)$ \tcp*{LLM parses constraints}
    \If{mandatory constraints missing}{
        \eIf{$ClarifyCounter < MaxClarify$}{
            $q \gets$ \textbf{ClarifyUser}$(q, Constraints)$ 
            Increment $ClarifyCounter$
        }{
            \textbf{break} \tcp*{Stop clarifying to avoid user fatigue}
        }
    }
}

$Candidates \gets$ \textbf{RetrieveModels}$(q)$ \tcp*{Embedding retrieval (Top K)}
$Filtered \gets$ \textbf{FilterCandidates}$(Candidates, Constraints)$ 

\If{$|Filtered| = 0$}{ 
    $BestMatch \gets$ \textbf{SelectClosestModel}$(Candidates, Constraints)$ \\
    $Explanation \gets$ \textbf{GenerateExplanation}$(q, BestMatch)$ \\
    \Return \{Recommendation: BestMatch, Explanation\}
}

\If{$|Filtered| > MaxCandidates$}{
    \If{$ClarifyCounter < MaxClarify$}{
        $q \gets$ \textbf{ClarifyUser}(q, Constraints) \\
        Increment $ClarifyCounter$ \\
        \textbf{Go to line 3} \tcp*{Restart process with clarified query}
    }
}

$Scores \gets$ \textbf{RankCandidates}$(q, Filtered)$ 
$OverallConfidence \gets$ \textbf{ComputeConfidence}$(Scores)$

\If{$OverallConfidence < ConfidenceThreshold$}{
    \If{$ClarifyCounter < MaxClarify$}{
        $q \gets$ \textbf{ClarifyUser}$(q, Constraints)$ \\
        Increment $ClarifyCounter$ \\
        \textbf{Go to line 3}
    }
}

$TopK \gets$ Top-$k$ candidates in $Filtered$ ranked by $Scores$


$Explanation \gets$ \textbf{GenerateExplanation}$(q, TopK)$

\Return \{Recommendations: TopK, Explanation\}
\end{algorithm}

The workflow of \system is shown in \cref{al: workflow}. The pipeline is implemented in Python using \texttt{pydantic} for schema validation, and the OpenAI GPT-based models for extraction. Each input document is processed in multiple iterations to collect diverse generations. The \fmd is stored in JSONL records and versioned via DVC to ensure reproducibility.

\section{LLM-Based In-Context Ranking Prompt}
\label{appendix:icl_prompt}

To re-rank candidate foundation models without training a dedicated learning-to-rank model, 
we leverage in-context learning~(ICL) with a LLM.
The prompt explicitly instructs the LLM to prioritize user requirements, compare candidate models, and produce a ranked list with explanations.
We provide few-shot examples created by an expert in the prompt to guide the model toward consistent ranking behavior. The prompt is connected to \fmd to provide the metadata of the candidate models. Below is the prompt template we are using in the ranking module:

\noindent
\textbf{Prompt Template:}
\begin{lstlisting}
You are an expert in remote sensing foundation model selection.

You will be given:
1. A structured user query specifying task requirements and constraints.
2. A list of candidate models retrieved from a database, each with metadata fields.

Your goal:
- Rank the candidate models from most to least suitable for the user's query.
- For each model, provide a brief explanation in several bullet points describing why it is placed at that rank.
- Prioritize hard constraints (application, modality, required sensor, and min_performance if provided), then consider secondary preferences (spatial/temporal resolution, application type, domain keywords, etc.).
- When two models equally satisfy the constraints and preferences, prefer the model that is more efficient, better validated on diverse benchmarks, or more versatile(multimodal, multi-temporal).

[Example]
Structured Query:
{
  "application": "land cover classification",
  "modality": "multispectral",
  "sensor": ["Sentinel-2"],
  "min_performance": {
    "metric": ["accuracy"],
    "value": [85]
  }
}

Candidate Models:
1. S2MAE
2. Prithvi
3. CACo

Ranking Output:
1. S2MAE
   - Directly supports Sentinel-2 multispectral data
   - Achieves 99.1\% accuracy on EuroSAT, exceeding 85\% requirement
   - Purpose-built for land cover classification
2. Prithvi
   - Supports multi-temporal multispectral data, including Sentinel-2
   - Accuracy slightly below requirement on similar tasks
   - More generalist FM
3. CACo
   - Only supports RGB modality
   - Accuracy below the 85\% requirement
   - Designed mainly for change detection and event retrieval

Your Task:
Given the following new query and candidates, produce a ranked list with explanations.

Structured Query:
{query}

Candidate Models:
{candidates}

Please output the ranked list as JSON in the following format:
[
  {
    "model": <model_name>,
    "rank": <integer>,
    "reason": [<short bullet points>]
  },
  ...
]
\end{lstlisting}

\section{Explanation Generator Prompt} 
\label{appendix:explanation_prompt}

The explanation generator uses an LLM to produce concise, interpretable justifications for the final ranked FM list. The prompt template in our explanation generator is given as follows:

\begin{lstlisting}
You are an expert in remote sensing foundation model selection.

The structured user query is:
{query}

The final ranked candidate models with their metadata are:
{ranked_models}

Your task:
1. For each model, output a JSON object with:
   - "model_name"
   - "explanation" (several bullet points on why it is recommended)
   - "paper_link"
   - "repository"
2. Highlight how the model satisfies or partially satisfies the query.
3. Mention key trade-offs if relevant (accuracy vs. efficiency, modality 
coverage, etc.).

\end{lstlisting}

\section{Prompt for RAG-LLM Baseline}
\label{appendix:rag_prompt}

For the \textsc{LLM-RAG} baseline, we prompt an LLM with the original user input and the retrieved model documentation as a context. 
The LLM is instructed to select and rank the top three remote sensing foundation models and provide concise explanations for each recommendation.

\begin{lstlisting}
You are an expert in remote sensing foundation models.

The user has provided the following task description:
{user_input}

Below is a set of candidate models with their documentation:
{context_str}

Your task: 
1. Select and rank the top 3 remote sensing foundation models most suitable for the task.
2. For each selected model, provide: 
-- A short explanation of why it fits the task requirements.
-- The reason for its ranking position compared to others.
-- Any other relevant information from the context.
3. Follow this exact output format: 

    1. model: <model_name> 
    explanation: 
    - <reason 1> 
    - <reason 2> 
    - <reason 3> 
    
    2. model: <model_name> 
    explanation: 
    - <reason 1> 
    - <reason 2> 
    - <reason 3> 
    
    3. model: <model_name> 
    explanation: 
    - <reason 1> 
    - <reason 2> 
    - <reason 3> 
\end{lstlisting}

\section{Expert Evaluation Procedure}
\label{appendix:expert_evaluation}

\stitle{Expert Background.}
All annotations were performed by two experts with a computer science background and specialization in RS. Both have prior experience working with RSFMs, have published in the relevant domains, and are familiar with model architectures, pretraining datasets, and evaluation practices.

\stitle{Annotation Protocol.}
To ensure consistency and reproducibility, we followed a structured, multi-stage scoring protocol:

\begin{itemize}[leftmargin=*]
    \item \textbf{Rubric Design.}  
    We created a detailed rubric for all seven criteria in \cref{tbl:criteria}, including definitions, examples, and decision rules.
    \item \textbf{Calibration Phase.}  
    Both experts annotated an initial subset of model-query pairs. Disagreements were used to refine the rubric until interpretations aligned.
    \item \textbf{Independent and Blind Scoring.}  
    Experts then rated all remaining model-query pairs independently and without access to system identities or each other's scores.
    \item \textbf{Disagreement Resolution.}  
    Any pair with substantial disagreement was re-examined in a controlled discussion, with decisions resolved strictly according to the rubric.
\end{itemize}

\stitle{Objective Scoring Rules.}
Where possible, we used explicit rules to reduce subjectivity:

\begin{itemize}[leftmargin=*]
    \item Reported Performance. Reported performance was determined by checking for benchmarks that matched the queried task. If none existed, we evaluated performance on broader but related tasks. For example, if the query specifies the task as scene classification, and there is no benchmark for this, we look for general classification benchmarks. Depending on its performance, this model gets a moderate/high reported performance score. Models with no relevant benchmarks received a low score.
    \item Efficiency. Model parameter counts were normalized to a 0-5 scale as a proxy for complexity, and combined with reported performance to obtain a final efficiency score. Specifically, we divide this complexity measure by the reported performance to produce a final efficiency score, also on a 0-5 scale. 
    Popularity. Popularity was used as a practical usability indicator rather than a measure of inherent model quality.  
    We used normalized GitHub star counts~(when code exists) and Google Scholar citation counts~(when paper is unavailable). This reflects maturity, community adoption, and available ecosystem support.
    \item Generalizability. We quantified pretraining diversity using three measurable components extracted from official FM documentation:
    \begin{enumerate}[leftmargin=*]
        \item Geographic diversity:  
        global (score 5), multi-regional (3--4), or single-region coverage (1--2).
        \item Sensor-modality diversity: 
        number of distinct modalities used in pretraining~\eg, optical, SAR, multispectral, hyperspectral).
        \item Dataset scale:  
        reported total area, number of scenes, or total images.
    \end{enumerate}
    These components were combined into a composite 1-5 score. Inter-annotator agreement confirmed that the rule-based definitions reduced subjectivity.
    \item Recency.
    Recency was defined by the publication year or the latest model-card update:
    \[
    \text{2025--2026}=5,\quad 2024=4,\quad 2023=3,\quad 2022=2,\ldots
    \]
    Given the rapid evolution in RSFMs, this criterion serves as a soft heuristic rather than a primary determinant.
\end{itemize}

\stitle{Reference Sources.}
All judgments were grounded in publicly available references for each foundation model. Experts used: (1) published papers and preprints; (2) official GitHub repositories and model documentation; (3) public benchmark results; (4) citation databases; and (5) described pretraining datasets from official sources. These references provided the necessary information on modality support, reported performance, efficiency, generalizability, popularity, and recency.

\section{Query Template for Creating Benchmark Dataset}
\label{appendix:templates}

\begin{table*}[t!]
\small
\caption{Structured query templates used for benchmark dataset generation. Each template maps to one constraint category. Slot values (\texttt{\{application\}}, \texttt{\{sensor\}}, \texttt{\{region\}}) are drawn from a predefined vocabulary and paraphrased by an LLM.}
\label{tbl:query_templates}
\centering
\begin{tabular}{p{0.83\textwidth} P{0.17\textwidth}}
\toprule
\textbf{Template} & \textbf{Categories} \\
\midrule
I’m looking for a model I can use out-of-the-box for \texttt{\{application\}} using \texttt{\{modality\}} data. I don’t have any labeled training data. & A1 \\
I have a well-labeled dataset for \texttt{\{application\}} with \texttt{\{modality\}} in \texttt{\{region\}}. Which model would be best to fully fine-tune from scratch? & A2 \\
I only have a few labeled samples for \texttt{\{application\}} using \texttt{\{sensor\}}. I want a model that can adapt well in a few-shot setting. & A3 \\
I have a lot of unlabeled \texttt{\{modality\}} imagery from \texttt{\{region\}}. I need a model that works well with self-supervised or unsupervised learning for \texttt{\{application\}}. & A4  \\
My data uses \texttt{\{sensor\}} with \texttt{\{spatial\_resolution\}} resolution, but most models I’ve seen don’t support it. Can you recommend one that can be adapted? & A5  \\
I'm working on \texttt{\{application\}} but only have access to a laptop with no GPU. Which model would be small enough to run locally? &  B1 \\
I’m using a desktop with a single GPU and doing \texttt{\{application\}} on \texttt{\{modality\}} imagery. Which models balance performance and efficiency? &  B2  \\
For \texttt{\{application\}}, I have access to cloud GPUs and can afford large models. What’s the most powerful foundation model I can try? & B3 \\
I'm doing basic \texttt{\{application\}} (\eg, 3–4 land classes). What lightweight model would you suggest for fast experimentation? &  C1 \\
I'm working on multi-class classification \texttt{\{application\}} with \texttt{\{modality\}} images. The task isn't trivial, but I don’t need pixel-level precision. &  C2  \\
I need a model for high-resolution segmentation or fine-grained \texttt{\{application\}}. Accuracy and spatial detail are important. & C3 \\
For \texttt{\{application\}} using \texttt{\{sensor\}} data, I mainly care about achieving the highest overall accuracy, even if the model is large. &D1\\
For \texttt{\{application\}} using \texttt{\{sensor\}} imagery, I want clean and accurate outputs with minimal false detections; clear boundaries and reliable predictions are most important. & D2\\
For \texttt{\{application\}} using \texttt{\{sensor\}} imagery, I need to ensure all target instances are captured, even if some false alarms occur; completeness is critical. &D3\\
I need fast inference for \texttt{\{application\}} in near real-time on \texttt{\{device\}}. What’s a good lightweight model? &  D4 \\
I'm doing \texttt{\{application\}} on \texttt{\{modality\}} in \texttt{\{region\}}, but I only have few-shot labels and limited compute. Which model fits this setup best? & Composite \\
\bottomrule
\end{tabular}
\end{table*}

To construct a representative and diverse benchmark dataset for evaluation, we define 16 structured query templates. Each template corresponds to a specific category of user constraints:

\begin{itemize}
    \item \textbf{Data Availability~(A1–A5):} 
    \begin{itemize}
        \item A1: \emph{No Training Data} — User wants to use pre-trained models directly.
        \item A2: \emph{Sufficient Labeled Data} — User has enough labels to fine-tune or train from scratch.
        \item A3: \emph{Few-shot Labels} — User has a small set of labeled data only and requires models that generalize in low-data regimes.
        \item A4: \emph{Unlabeled Data Only} — User has input data but no labels and seeks models suited for unsupervised or self-supervised settings.
        \item A5: \emph{Data Adaptation Needed} — User's data differs from typical inputs, requiring domain adaptation or compatibility adjustments.
    \end{itemize}
    
    \item \textbf{Computational Resources~(B1–B3):}
    \begin{itemize}
        \item B1: \emph{Limited Resources} — \eg, CPU-only laptop.
        \item B2: \emph{Moderate Resources} — \eg, desktop with GPU.
        \item B3: \emph{High Resources} — \eg, cluster-scale GPU compute.
    \end{itemize}
    
    \item \textbf{Application Complexity~(C1–C3):}
    \begin{itemize}
        \item C1: \emph{Simple Application} — Applications with low label granularity or few classes~(\eg, binary classification, basic change detection).
        \item C2: \emph{Moderate Application} — Applications with moderate difficulty, such as multi-class classification or coarse semantic segmentation.
        \item C3: \emph{Complex Application} — Applications requiring fine-grained spatial precision, multi-class segmentation, multi-modal fusion, or high-resolution outputs.
    \end{itemize}

    \item \textbf{Evaluation Priorities~(D1–D4):}
    \begin{itemize}
        \item D1: \emph{Accuracy-Focused} — Maximize correctness of classification or segmentation outcomes.
        \item D2: \emph{Output Quality-Critical} — Prioritize clean, well-bounded, and visually reliable outputs~(\eg, high mIoU, sharp edges, no artifacts).
        \item D3: \emph{Coverage-Critical} — Ensure all relevant regions or objects are detected, even at the cost of some false positives~(\eg, disaster mapping, change detection).
        \item D4: \emph{Speed-Critical} — Require lightweight or low-latency models for fast inference on edge devices.
    \end{itemize}
\end{itemize}

Accordingly, \cref{tbl:query_templates} shows the full list of templates used to generate the benchmark queries. Slot values~(\eg, \texttt{\{application\}}, \texttt{\{sensor\}}, \texttt{\{region\}}) are drawn from a predefined vocabulary and instantiated using sampling and LLM-based paraphrasing.

\section{Expert Scoring Weight Configuration}
\label{appendix:weights}

To aggregate model evaluation scores during expert labeling, we apply a weighted linear combination of the seven criteria from \cref{tbl:criteria}. The weights are as follows:

\begin{table}[H]
  \centering
  \small
\begin{tabular}{lc}
\toprule
\textbf{Criterion} & \textbf{Weight (\%)} \\
\midrule
Application Compatibility & 25 \\
Modality Match & 20 \\
Reported Performance & 20 \\
Efficiency & 15 \\
Generalizability & 10 \\
Popularity & 5 \\
Recency & 5 \\
\bottomrule
\end{tabular}
\end{table}

These weights were empirically determined on the basis of expert interviews. We normalize raw scores before aggregation.

\section{Illustrative Examples of Expert Scoring}
\label{appendix:expert_examples}

To improve transparency, we provide several examples demonstrating how experts applied the scoring rubric to real model-query pairs.  
Each example includes:  
(1) the natural-language query,  
(2) the top-3 FM selections from all systems, and  
(3) the expert ratings across the seven criteria defined in \cref{tbl:criteria}.  
These examples show how rubric-guided, independent scoring yields consistent and interpretable evaluations.

\etitle{Example 1:}

\textbf{Query:}  
\emph{I need a model for fine-grained land cover classification using high-resolution multispectral imagery. Accuracy and spatial detail are important.}

\textbf{Selected FMs~(Top-3 from Each System):} See \cref{tbl:evaluation_results_compact}.

\etitle{Example 2:}

\textbf{Query:}  
\emph{I only have a few labeled samples for urban expansion detection using Sentinel-1 and Sentinel-2 time series data from 2016-2023. I want a model that can adapt well in a few-shot setting.}

\textbf{Selected FMs~(Top-3 from Each System):} See \cref{tbl:evaluation_results_compact}.

\begin{table}[t]
\renewcommand\arraystretch{0.8}
\centering
\scriptsize
\caption{
Evaluation results for queries 1 and 2.  
\textbf{Criteria:}  
CR1 - Application Compatibility;  
CR2 - Modality Match;  
CR3 - Reported Performance;  
CR4 - Efficiency;  
CR5 - Generalizability;  
CR6 - Popularity;  
CR7 - Recency.}
\label{tbl:evaluation_results_compact}

\setlength{\tabcolsep}{2pt}

\begin{tabular}{
    m{2cm}  
    c         
    c         
    c c c c c c c c
}
\toprule
\textbf{System} & \textbf{Rank} & \textbf{FM} & \textbf{CR1} & \textbf{CR2} & \textbf{CR3} &
\textbf{CR4} & \textbf{CR5} & \textbf{CR6} &
\textbf{CR7} & \textbf{Final Score} \\
\midrule

\multicolumn{11}{c}{\textbf{Query 1}} \\[2pt]
\midrule
\multirow{3}{*}{\system}
  & 1 & OmniSat       & 5 & 5 & 5   & 5   & 4   & 3   & 4 & 94 \\
  & 2 & FlexiMo       & 4 & 4.5 & 4 & 2.5 & 1.5 & 3.5 & 5 & 75 \\
  & 3 & CtxMIM        & 5 & 5 & 4.5 & 3 & 1.5 & 3.5 & 3 & 83.5 \\
\cmidrule(lr){1-11}

\multirow{3}{*}{\system-Naive}
  & 1 & OmniSat       & 5 & 5 & 5 & 5 & 4 & 3 & 4 & 94 \\
  & 2 & FlexiMo       & 4 & 4.5 & 4 & 2.5 & 1.5 & 3.5 & 5 & 75 \\
  & 3 & CtxMIM        & 5 & 5 & 4.5 & 3 & 1.5 & 3.5 & 3 & 83.5 \\
\cmidrule(lr){1-11}

\multirow{3}{*}{DB-Retrieval}
  & 1 & SpectralEarth & 3 & 3 & 3.5 & 1.5 & 3 & 3 & 5 & 59.5 \\
  & 2 & OmniSat       & 5 & 5 & 5   & 5   & 4   & 3   & 4 & 94 \\
  & 3 & MATTER        & 4 & 4.5 & 4 & 4.5 & 3.5 & 1 & 2 & 75 \\
\cmidrule(lr){1-11}

\multirow{3}{*}{Unstr.-RAG}
  & 1 & FoMo          & 5 & 5 & 3.5 & 1.5 & 2   & 1.5 & 5 & 79.5 \\
  & 2 & DynamicVis    & 4 & 4 & 4   & 3.5 & 3.5 & 2   & 5 & 75 \\
  & 3 & SatVision-TOA & 2.5 & 4 & 2.5 & 0 & 2.5 & 5 & 4 & 55 \\

\midrule

\multicolumn{11}{c}{\textbf{Query 2}} \\[2pt]
\midrule
\multirow{3}{*}{\system}
  & 1 & SSL4EO-S12      & 5 & 5 & 4 & 4 & 4.5 & 4.5 & 3 & 89.5 \\
  & 2 & Ial-SimCLR      & 3.5 & 5 & 3.5 & 5 & 2 & 3 & 3 & 77.5 \\
  & 3 & SeCo            & 3 & 3 & 3.5 & 5 & 5 & 2.5 & 1 & 67 \\
\cmidrule(lr){1-11}

\multirow{3}{*}{\system-Naive}
  & 1 & SoftCon         & 5 & 5 & 4.5 & 3 & 3 & 4 & 4 & 87 \\
  & 2 & SkySense        & 5 & 5 & 5   & 1 & 3.5 & 5 & 4 & 85.5 \\
  & 3 & SSL4EO-S12      & 5 & 5 & 4 & 4 & 4.5 & 4.5 & 3 & 89.5 \\
\cmidrule(lr){1-11}

\multirow{3}{*}{DB-Retrieval}
  & 1 & CACo            & 3 & 3 & 4 & 4 & 4 & 4 & 3 & 70 \\
  & 2 & SeCo            & 3 & 3.5 & 5 & 5 & 5 & 2.5 & 1 & 67 \\
  & 3 & SSL4EO-S12      & 5 & 5 & 4 & 4 & 4.5 & 4.5 & 3 & 89.5 \\
\cmidrule(lr){1-11}

\multirow{3}{*}{Unstr.-RAG}
  & 1 & CACo            & 3 & 3 & 4 & 4 & 4 & 4 & 3 & 70 \\
  & 2 & Copernicus-FM   & 3 & 3.5 & 3 & 1 & 3.5 & 5 & 5 & 62.5 \\
  & 3 & AnySat          & 3.5 & 5 & 3.5 & 1.5 & 4 & 4.5 & 5 & 74 \\
\bottomrule

\end{tabular}

\end{table}

These examples illustrate how the rubric was applied in practice and how expert judgments reflect both task requirements and model capabilities.  
They also demonstrate how rubric-guided scoring minimizes subjective variation across annotators.





\end{document}